\documentclass[conference]{IEEEtran}
\IEEEoverridecommandlockouts
\usepackage{cite}
\usepackage{amsmath,amssymb,amsfonts}
\usepackage{algorithm}
\usepackage{algpseudocode}
\usepackage{graphicx}
\usepackage{multirow}
\usepackage{textcomp}
\usepackage{xcolor}

\def\BibTeX{{\rm B\kern-.05em{\sc i\kern-.025em b}\kern-.08em
    T\kern-.1667em\lower.7ex\hbox{E}\kern-.125emX}}
\begin{document}

\title{Synaptic metaplasticity with multi-level memristive devices\\
\vspace{-0.8cm}
}


\author{
\IEEEauthorblockN{
S. D'Agostino\textsuperscript{*},
F. Moro\textsuperscript{*}, 
T. Hirtzlin\textsuperscript{*},
J. Arcamone\textsuperscript{*}, 
N. Castellani\textsuperscript{*},
D. Querlioz\textsuperscript{$\dagger$},
M. Payvand\textsuperscript{$\ddagger$} and
E. Vianello\textsuperscript{*}}
\IEEEauthorblockA{
Email: \{simone.dagostino, elisa.vianello\}@cea.fr \\
\textsuperscript{*}\textit{\ CEA-Leti, Université Grenoble Alpes, F-38000 Grenoble, France}\\ 
\textsuperscript{$\dagger$}\textit{\ Université Paris-Saclay, CNRS, Centre de Nanosciences et de Nanotechnologies, Palaiseau, France }\\
\textsuperscript{$\ddagger$}\textit{\ Institute for Neuroinformatics, University of Zurich and ETH Zurich, Zurich, Switzerland}
\vspace{-0.55cm}}
}

\maketitle

\begin{abstract}
Deep learning has made remarkable progress in various tasks, surpassing human performance in some cases. However, one drawback of neural networks is catastrophic forgetting, where a network trained on one task forgets the solution when learning a new one. To address this issue, recent works have proposed solutions based on Binarized Neural Networks (BNNs) incorporating metaplasticity. In this work, we extend this solution to quantized neural networks (QNNs) and present a memristor-based hardware solution for implementing metaplasticity during both inference and training.
We propose a hardware architecture that integrates quantized weights in memristor devices programmed in an analog multi-level fashion with a digital processing unit for high-precision metaplastic storage.
We validated our approach using a combined software framework and memristor based crossbar array for in-memory computing fabricated in 130~nm CMOS technology.
Our experimental results show that a two-layer perceptron achieves $\mathbf{97}\,$\% and $\mathbf{86}\,$\% accuracy on consecutive training of MNIST and Fashion-MNIST, equal to software baseline.
This result demonstrates immunity to catastrophic forgetting and the resilience to analog device imperfections of the proposed solution. Moreover, our architecture is compatible with the memristor limited endurance and has a $\mathbf{15\times}$ reduction in memory footprint compared to the binarized neural network case.

\vspace{0.25cm}
\end{abstract}
\section{Introduction}

\begin{figure}[ht]
    \centering
    \includegraphics[width=0.485\textwidth]{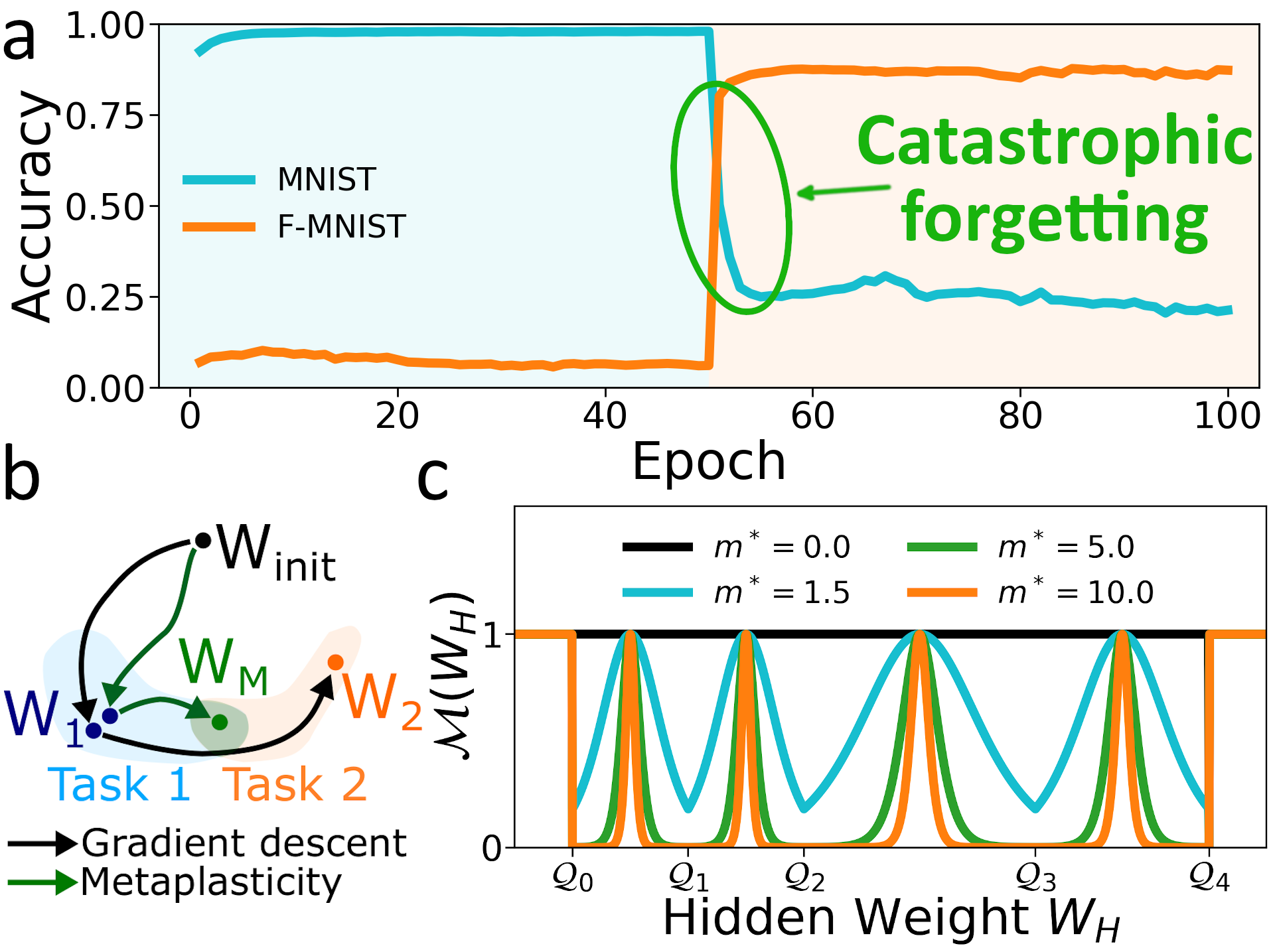}
    \caption{\textbf{a} The ``catastrophic forgetting'' problem: a network is trained sequentially with two different training sets (here MNIST and Fashion-MNIST). When learning Fashion-MNIST, the MNIST test accuracy collapses almost to random guessing. \textbf{b} The black arrows depict the paths inside the parameter space when using a classic learning sequence of MNIST and Fashion-MNIST, while the green arrows show the paths traversed using the mataplasticity training. \textbf{c} Metaplastic function $\mathcal{M}$ on a set of unequally spaced $\mathcal{Q}$ levels; it is used to implement metaplasticity on QNN by modulating the weights updates.}
    \label{fig:f1}
\end{figure}

Intelligence in mammals is characterized by the ability to learn, which encompasses both the acquisition of new knowledge and skills, as well as the retention of previously acquired information. In contrast, state-of-the-art deep neural networks suffer from ``catastrophic forgetting'' (Fig.~\ref{fig:f1}a), where the network forgets previously learned information when learning new information \cite{m1}. Recent advancements in the areas of class incremental learning \cite{m2}, meta-learning \cite{m3}, and metaplasticity \cite{m4, m5} have shown promise in addressing this problem. When learning a new task, the synaptic weights optimized during previous tasks are protected from further updates, allowing the network to find a set of parameters that can solve both tasks simultaneously (Fig.~\ref{fig:f1}b).
These algorithms can enable continuous learning from the sensory signals which has applications in tailoring the edge devices to unique users. For example, customizing wearable medical devices to suit each patient's needs, or adjusting smart home devices to reflect users’ habits or preferences. Implementing such continuous learning algorithms on the chip will remove the need for sending the user/patient’s data to the cloud, which not only saves significant amounts of power consumption but also guarantees their privacy.
On-chip learning can benefit greatly from on-chip, high-density, and analog memory~\cite{m6}, and resistive memory technologies have emerged as a good candidate solution~\cite{m7, m8, m9}. 
Despite its great potential, resistive memory undergoes variability and, in practice, the bit resolution is usually limited to up to 3 bits~\cite{m10}. \\
\indent In this work, we propose a device-algorithm co-design approach that takes advantage of the recent meta-plastic algorithms implemented for Binary Neural Networks (BNNs)~\cite{m5}, while being compatible with resistive memory characteristics. \\
\indent Specifically, we first demonstrate that the metaplasticity-inspired training method for BNNs can be extended to quantized neural networks (QNNs) with more than two binary states. 
By implementing quantization in hardware, we are able to capitalize on the multi-level capabilities of memristors in a crossbar array to perform analog matrix-vector multiplication operations with low latency and energy consumption. Furthermore, we propose a mixed-precision architecture that combines the use of a memristor crossbar array for storing synaptic weights and performing matrix-vector multiplication operations with a digital processing unit for storing metaplastic variables in high precision and calculating low-precision weight updates based on these values~\cite{m7}. To validate the effectiveness of our metaplasticity-inspired training method on the mixed-precision architecture, we conduct a combined hardware/software training experiment using a recently developed memristor crossbar array for in-memory computing~\cite{m10}. Our key contributions include:
\begin{itemize}
    \item a metaplasticity rule inspired by \cite{m5} for quantized neural networks that reduces catastrophic forgetting,
    \item experimental validation on a 16~kbit crossbar, showing that each 1T1R cell can store nine conductance levels per memristor device,
    \item an online implementation of the proposed training technique on memristor-based hardware,
    \item a study of this approach with respect to the limited compute precision and imperfection of the memristor devices.
\end{itemize}

\section{Synaptic Metaplasticity in Quantized Neural Networks}
QNNs are a generalization of BNNs that use multiple levels of quantization \cite{m11}. In QNNs, synapses consist of two types of weights: quantized weights, $W_S$, and hidden weights, $W_H$. The hidden weights are arbitrary precision real values, while the quantized wights are discretized to a set of values, $\mathcal{Q}=\{\mathcal{Q}_i\}$ with $i=1, ..,n$. The training process for QNNs involves updating the hidden weights using loss gradients computed with the quantized weights (as outlined in Algorithm 1, lines 2-3). The quantized weights are chosen as the closest quantized values to the hidden weights:

\begin{equation}
    W_S=\mathcal{Q}_i:|W_S-W_H|=\min_{i=0}^n\left\{|\mathcal{Q}_i-W_H|\right\}
    \label{eqn:1}
\end{equation}

In \cite{m5}, it has been proposed that the hidden weights in BNNs may be interpreted as metaplastic states by introducing an additional component into the hidden weight update rule. This component allows for a memory effect by introducing a meta-function, $\mathcal{M}(W_H)$. The $\mathcal{M}(W_H)$ function modulates the strength of the updates: when $\mathcal{M}(W_H)$ is low, the hidden weight is consolidated at its current level, while when it is high it is updated according to the stochastic gradient descent learning algorithm (lines 7-10 of Algorithm 1).
The meta-function $\mathcal{M}(W_H)$ can be extended to QNNs using the following assumption: $\mathcal{M}(W_H)$ decreases as it approaches a quantized level and reaches a maximum at the midpoint between two adjacent quantized levels. In the interval between two adjacent quantized levels, $\mathcal{I}^D_i=\mathcal{Q}_{i+1}-\mathcal{Q}_i$, the metaplastic function is defined as a function of the hidden weight value: 
\begin{equation}
\mathcal{M}(W_H) = 1-\tanh^2\left(\dfrac{2m^*}{\mathcal{I}^D_i}|W_H-W_S|-m^*\right)
\label{eqn:2}
\end{equation}
with $W_H\in \mathcal{I}^D_i$ and $W_S$ the quantized weight (Eq.~\ref{eqn:1}).
The $m^*$ parameter controls the steepness of the decay and therefore determines the rate of the consolidation of the hidden weight. An example of $\mathcal{M}(W_H)$ function for four quantized levels and various values of the scalar $m^*$ is shown in Fig.~\ref{fig:f1}c. 

In this work, all simulations use adaptive moment estimation (Adam) \cite{m12}. The training procedure for the QNN with metaplasticity is outlined in Algorithm 1. In this algorithm, $\mathbf{W_H}$ represents the vector of hidden weights, with $W_H$ representing a single component. A similar notation is used for the other vectors. The batch-norm parameters are represented by $\mathbf{\theta^{BN}}$, while $\mathbf{U_W}$ and $\mathbf{U_\theta}$ represent the updates to the weights and the batch-norm, respectively. The input and target vectors are represented by $(\mathbf{x}, \mathbf{y})$, and the learning rate is represented by $\eta$.

\begin{algorithm}
    \caption{Algorithm for metaplasticity using multiple quantized levels.}
    \begin{algorithmic}[1]
    \Require $\mathbf{W_H, \theta^{BN}, U_W, U_{\theta}, (x, y)}, m^*, \eta, \mathbf{\mathcal{Q}}$
    \Ensure $\mathbf{W_H, \theta^{BN}, U_W, U_{\theta}}$
        \State $\mathbf{W_S} \gets \texttt{Approx}(\mathbf{W_H}, \mathbf{\mathcal{Q}})$  \Comment{Eq.~\ref{eqn:1}}
        
        \State $\mathbf{\hat{y}} \gets \texttt{Forward}(\mathbf{x}, \mathbf{W_S}, \mathbf{\theta^{BN}})$
        
        \State $\mathbf{C} \gets \texttt{Cost}( \mathbf{\hat{y}}, \mathbf{y})$
        
        \State $\partial_{\mathbf{W}}C, \partial_{\mathbf{\theta}}C \gets \texttt{Backward}(C, \mathbf{\hat{y}}, W_S, \mathbf{\theta^{BN}})$
        
        \State $\mathbf{U_W}, \mathbf{U_{\theta}} \gets \texttt{Adam}(\partial_WC, \partial_{\mathbf{\theta}}C, \mathbf{U_W}, \mathbf{U_{\theta}})$
        \For{$W_H$ in $\mathbf{W_H}$}
        \If{$U_W\cdot(W_H-W_S)<0$}
        \State $W_H \gets W_H - \eta U_W\mathcal{M}(m^*,W_H,W_S)$  \Comment{Eq.~\ref{eqn:2}}
        \Else
        \State $W_H\gets W_H-\eta U_W$
        \EndIf
        \EndFor
    \end{algorithmic}
    \label{alg:a1}
\end{algorithm}

\begin{figure}
    \centering
    \includegraphics[width=0.485\textwidth]{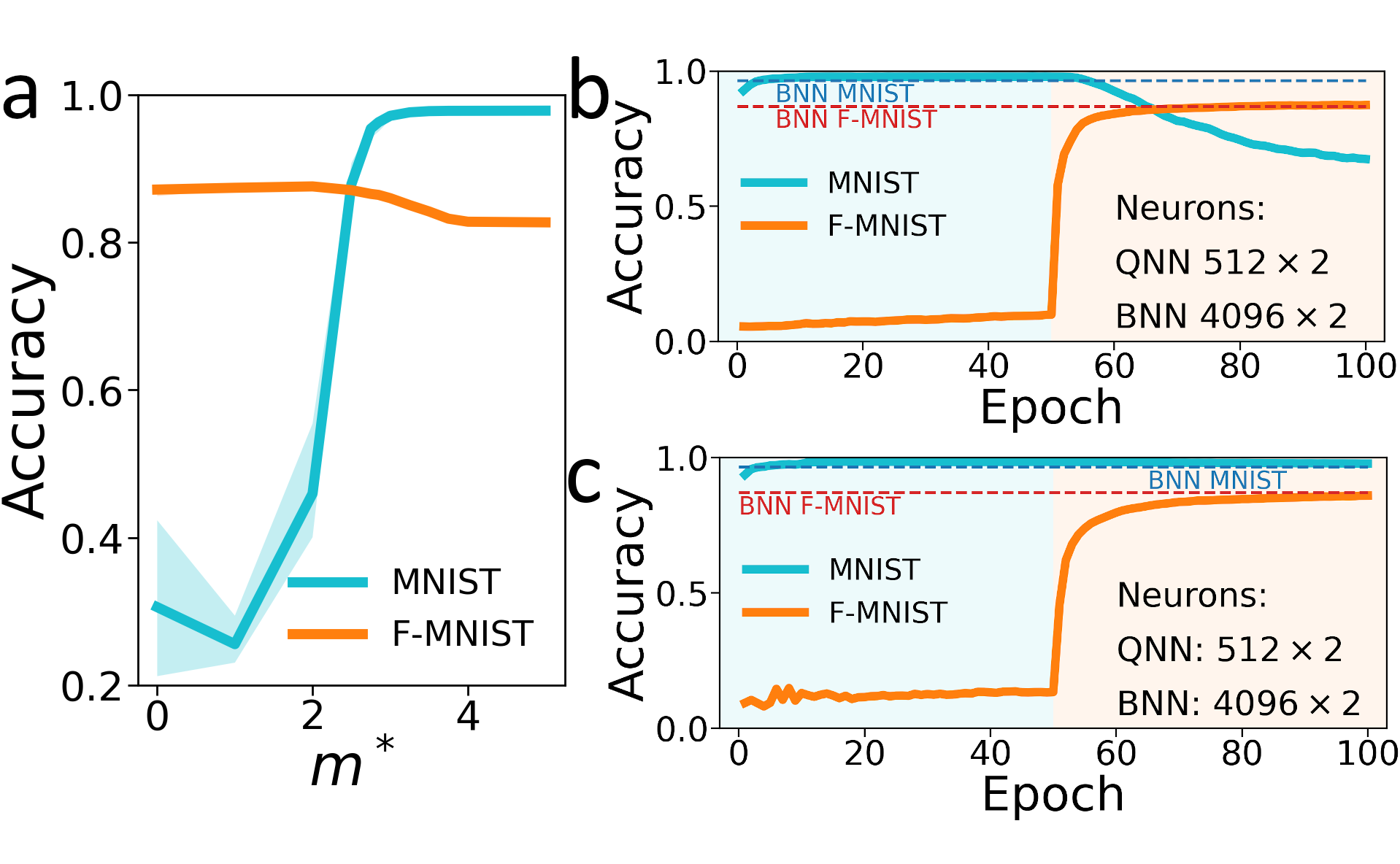}
    \caption{\textbf{a} Impact of the $m^*$ factor on the MNIST and Fashion-MNIST sequential learning: accuracy plot after $50$ epochs. \textbf{b} Sequential learning of MNIST and Fashion-MNIST with $m^*=2.2$. Comparison with a BNN with two hidden layers of $4096\,$neurons each. \textbf{c} Sequential learning of MNIST and Fashion-MNIST with $m^*=3$. Baseline: BNN with two hidden layers of $4096\,$neurons each.}
    \label{fig:f2}
\end{figure}

The algorithm has been tested using a multi-layer perceptron (MLP) with two hidden layers of 512 neurons each, trained on the MNIST dataset\cite{m13} first (epochs 1-50) and then on Fashion-MNIST\cite{m14} (epochs 51-100). The levels $\mathcal{Q}$ are $17$ in the range $[-1.5, 1.5]$. The training experiment was performed using $m^*$ values ranging from 0 to 5. To ensure proper weight initialization, the first $10\,$epochs were preformed without metaplasticity ($m^*=0$). For $m^*<2$, the neural network experienced catastrophic forgetting, but in the interval $m^*\in[2, 4]$, the network was able to learn both tasks with near-independent task accuracy. However, when $m^*>4$, the consolidation rate is too strong and the accuracy on 
Fashion-MNIST drops without any significant gain from the lower forgetting rate on MNIST (Fig.~\ref{fig:f2}).


The results show that using $m^*=3$ leads to an accuracy of over $97\,$\% for MNIST and $86\,$\% for Fashion-MNIST. This value of $m^*=3$ is applied through the rest of the article. These results match those obtained using a two-layer BNN with $4096\,$neurons per layer in previous studies \cite{m5}.


\section{Memristor-based implementation}
\begin{figure}
    \centering
    \includegraphics[width=0.485\textwidth]{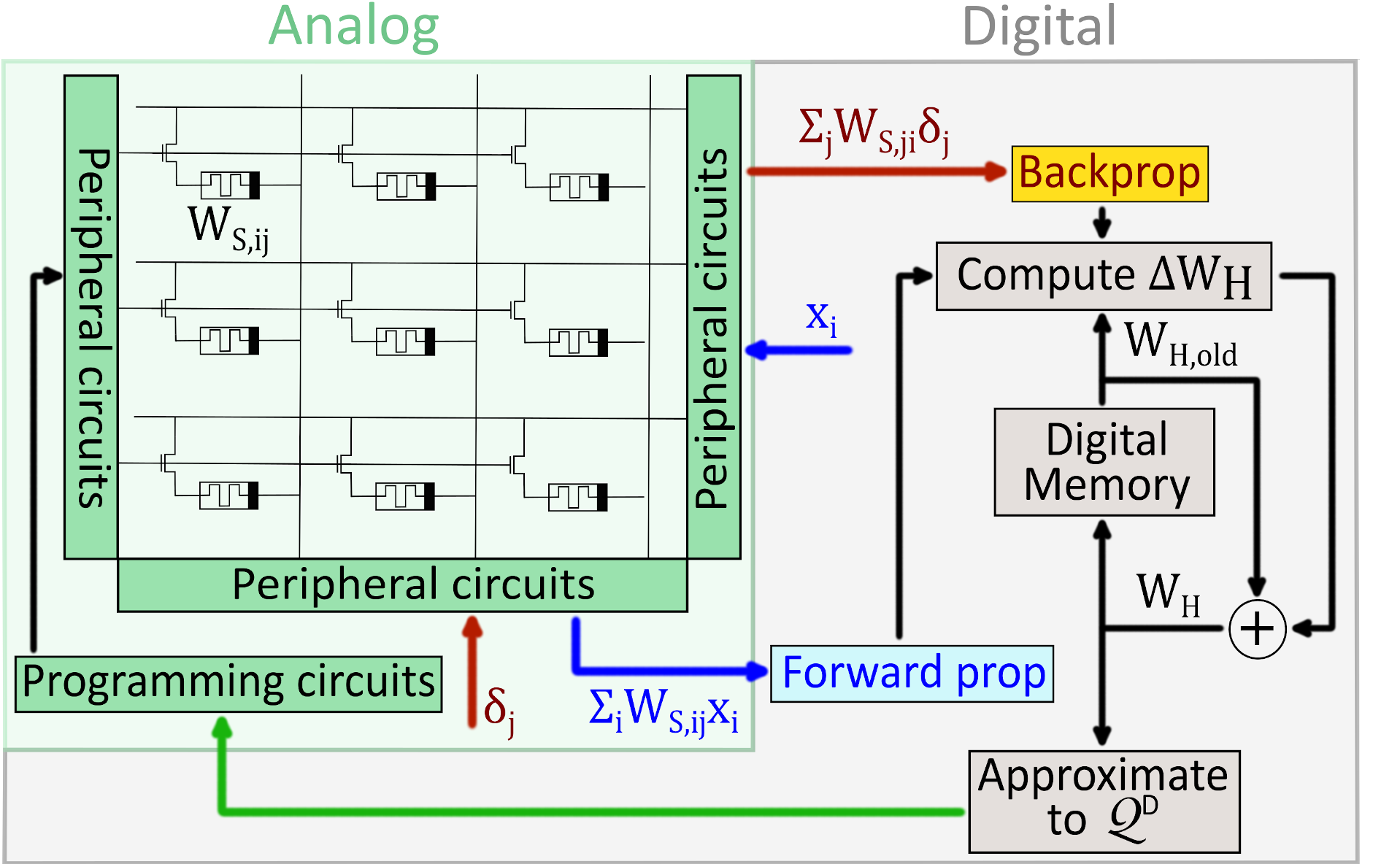}
    \caption{Schematic of the on-chip mixed analog/digital learning architecture. The analog in-memory computing block has several crossbar arrays programmed in a multi-level fashion to store the hidden wights. This analog block performs the \textit{Forward} (blue) and \textit{Backward} (red) propagations. The resulting values are used to compute the hidden weights updates stored in the digital memory. The memristors conductance values are updated accordingly.}
    \label{fig:f3}
    \vspace{-1pt}
\end{figure}

The proposed system (Fig.~\ref{fig:f3}) for training QNNs has two components: a memristive crossbar array for analog in-memory computing (green box) and a high-precision digital computational unit (grey box). Memristors can store intermediate conductance levels, unlike CMOS-based memories (SRAM or DRAM) which store one bit per cell. This allows quantized weights in a QNN to be directly stored as conductance levels in a memristor crossbar array, resulting in compact weight storage.
Additionally, the multiply and accumulate operations necessary during the forward and backward data propagation stages of QNNs training can be performed in-place using the fundamental laws of electric circuits: the multiply operation corresponds to Ohm's law, while the accumulate one to Kirchoff's current law. During forward propagation, the neuron activations ($x_i$) are transmitted to the Source Lines through voltages. The total current flowing through each column is the sum of the product of the weights ($W_{S,ij}$) and the activations ($x_i$) along each Bit Line. In backward propagation, the errors, $\delta_i$, are applied to the Bit Lines and the resulting currents $I=\sum_jW_{S,ji}\delta_j$ measured at the Source Lines. These current values are used to compute the gradient respect to the quantized weights stored in the analog crossbar arrays. The desired hidden weight updates are then calculated based on the the gradient calculated on the quantized weights and the metaplasticity rule (Algorithm 1, lines 5-11). The new hidden weights $W_H$ are updated in the high-precision digital memory. At last, the hidden weights are approximated to the available levels $\mathcal{Q}$ and the memristors in the crossbar are eventually re-programmed by a dedicated programming circuit, if the quantized level has changed.

\begin{figure}
    \centering
    \includegraphics[width=0.485\textwidth]{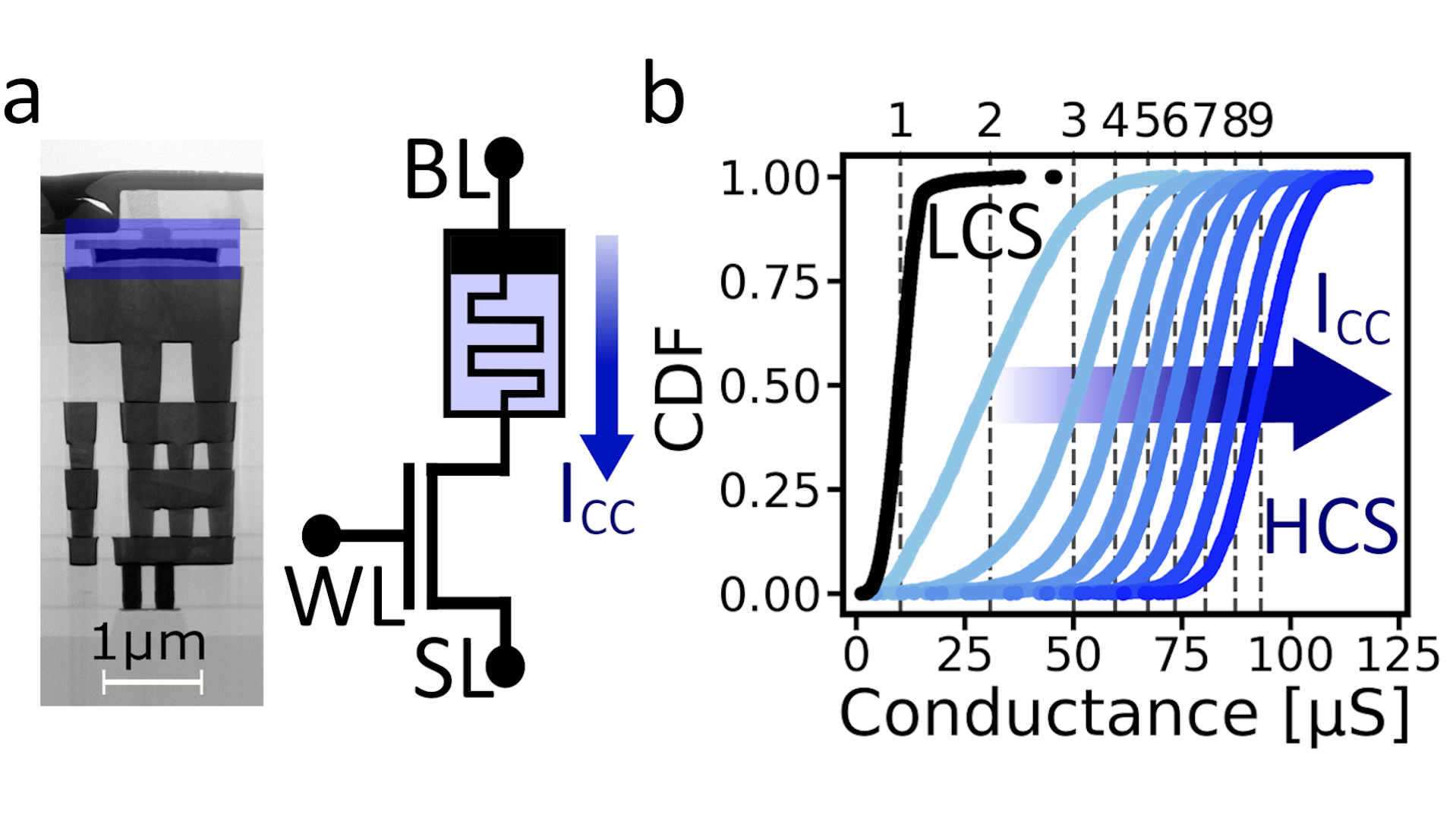}
    \caption{\textbf{a} 1T1R hafnium-based memristive device in a SEM image with highlighted memristor (blue). \textbf{b} Cumulative Density Function of 8-Level HCS, programmed with different $I_{CC}$ programming current values, and LCS.}
    \label{fig:f4}
\end{figure}

To validate the feasibility of the proposed architecture, we fully characterized an analog in-memory-computing circuit in hybrid CMOS/memristor process \cite{m15}. Hafnium oxide (HfO\textsubscript{X}) memristors are fabricated on top of a CMOS foundry $130\,$nm process with four levels of metals. Fig.~\ref{fig:f4}a shows a Scanning Electron Microscope (SEM) image of a fabricated 1-transistor 1-resistor (1T1R) memory cell. The memristor device starts in an extremely low conductance state (pristine state) when manufactured. The device must experience a unique ``forming'' operation to create a first conductive filament. Once formed, memristors can be programmed to multiple conductance levels. A set operation programs the device in the High-Conductance-State (HCS), a reset operation sets it into the Low-Conductance-State (LCS). The conductance value is controlled  by modulating the set programming current $I_{CC}$, defined by the gate voltage applied on the selector transistor~\cite{m16}. We characterized a $16\,$kb array of 1T1R devices, programming them with eight different compliance currents (HCS modulated by $I_{CC}$) and the LCS state, the corresponding conductance distributions are show in Fig.~\ref{fig:f4}b. However, memristors are prone to a large conductance variability, resulting in a broad statistical distribution of conductance values after programming. A second challenge for the proposed architecture is the limited endurance (i.e. number of Set-Reset operations) of the devices.

\begin{figure}
    \centering
    \includegraphics[width=0.485\textwidth]{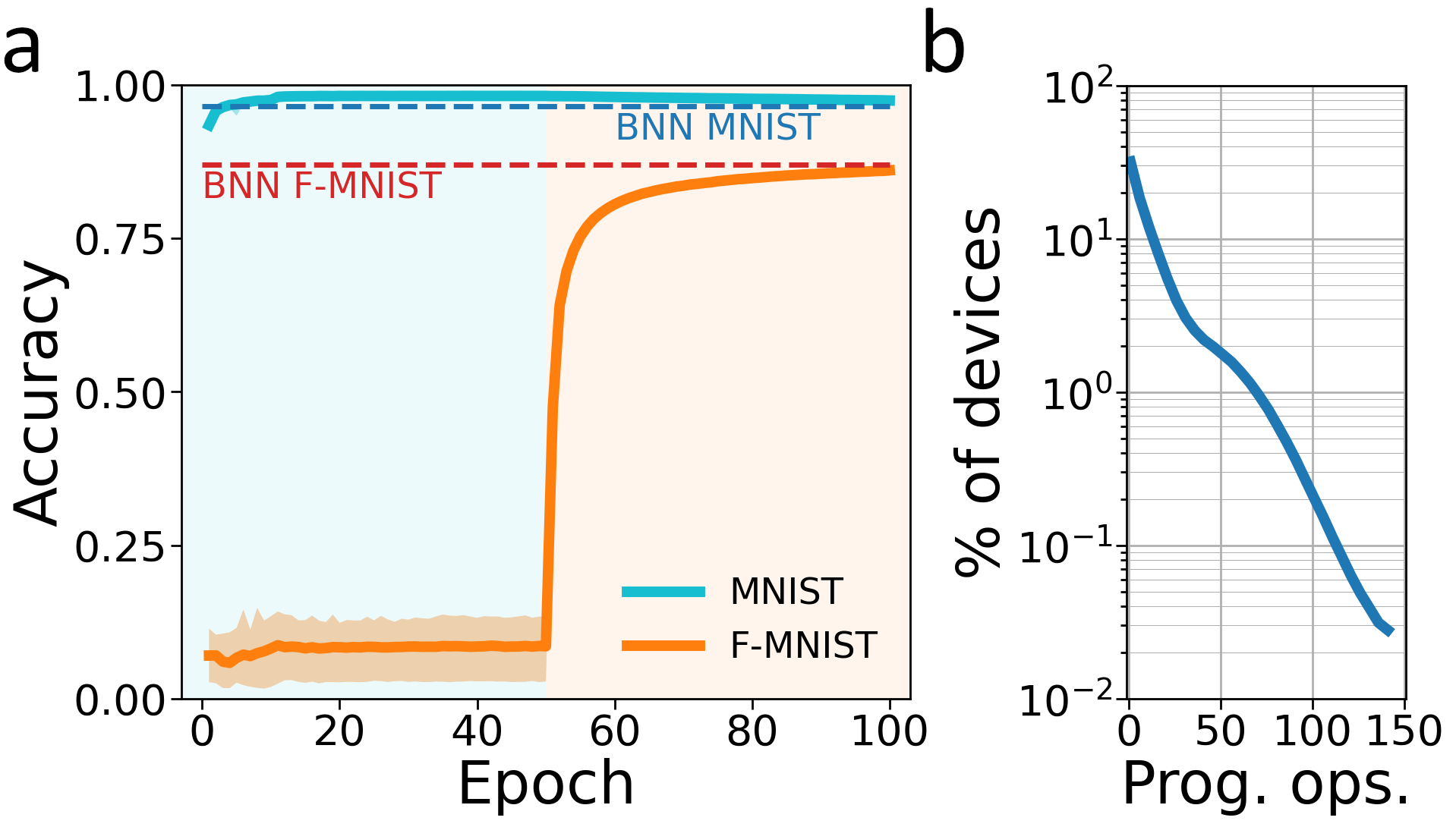}
    \caption{\textbf{a} Sequential learning of MNIST and Fashion-MNIST in hybrid software/hardware experiment with 15 repetitions. Comparison with a BNN with two hidden layers of $4096\,$neurons each. \textbf{b} Percentage of devices as a function of the number of programming operations after training on MNIST and Fashion-MNIST.}
    \label{fig:f5}
\end{figure}

We evaluated the efficacy of our proposed architecture through a series of experiments and simulations. We trained a two layer-perceptron on MNIST for $50$ epochs, using the first $10$ epochs as a pre-training phase (i.e. no metaplasticity, $m^*=0$) before switching to Fashion-MNIST for the next $50$ epochs. We utilized the same network architecture discussed in the previous Section. Quantized weights are encoded as the difference in conductance between two adjacent memory cells, enabling storage of both positive and negative weights \cite{m15}. Since each memristor can be programmed into nine levels (8 HCS levels and 1 LCS level), each weight can be represented by 17 levels, with the ``zero'' level obtained by matching the conductance of two memristors in two adjacent cells. The initial values for the hidden weights are determined by drawing samples from Gaussian distributions with a mean value set between two adjacent quantized levels. These quantized weights are then programmed into conductance values using the equation Eq.~(\ref{alg:a1}). These conductance values are read from the hardware and used in the simulation stage to compute the hidden weights updates through software simulations. The conductance values are then updated as necessary, with single-shot pulses applied without adjusting for the difference between the desired and observed conductance change. At the end of each training epoch, the conductance values for all memristors are read from the array and used to evaluate the classification performance. The results, shown in Fig.~\ref{fig:f5}a, indicate that the memristor conductance variability does not significantly affect accuracy. This is comparable to other works that showed as the noise added during training should not affect or, in some cases, enhance the performances of QNNs\cite{m17, m18, m19}. 
The network achieves a maximum accuracy of $97.47\pm0.33,$\%  for MNIST and $86.09\pm0.33,$\% for Fashion-MNIST. This suggests that metaplasticity in QNNs is robust to memristor conductance variability during training, making on-chip learning a viable option.\\
\indent In addition to evaluating the accuracy, we computed the number of programming operations required throughout the experiment while learning both tasks. We plot the percentage of devices in the network against the number of programming operations they have undergone in Fig.~\ref{fig:f5}b. The figure shows that the majority of devices ($76.12\,$\%) only require less than $25$ operations, with only a small number of memristors ($10.16\,$\%) requiring more than $50$ programming operations. This is orders of magnitude lower than the endurance of memristors, which has been measured to be around $10^5$ programming cycles \cite{m20}. The low number of programming operations results from updating only hidden weights at each iteration of the training algorithm, while re-programming memristors only if the associated quantized level changes.\\ 
\indent It is worth noting that our model has $1.3\cdot10^6$ devices for implementing quantized weights and reaches similar accuracy on MNIST and Fashion-MNIST compared to a BNN with $20\cdot10^6$ binary weights. In terms of memristor-based implementations, the proposed algorithm allows for a reduction in memory footprint by a factor of $15\times$ and $30\times$ compared to the binarized-neural-network implemented using a one-memristor one-transistor cell (1T1R) \cite{m21} and a two-memristors two-transistors cell (2T2R) \cite{m22}, respectively (Table~\ref{tab:1}).

\begin{table}[ht]
    \caption{Comparison of results between the BNN network and QNNs of different sizes including OxRAM variability}
    \vspace{-0.3cm}
    \begin{center}
    \def\arraystretch{1.25}
    \begin{tabular}{|c|c|c|c|c|}
    \hline
       &\textbf{BNN\cite{m5}}&  \multicolumn{3}{c|}{\textbf{QNN hardware implementation}}\\ 
       \hline
       Number & $20\,$M$^{\mathrm{a}}$ &\multirow{2}{*}{$537\,$k$^{\mathrm{b}}$}  & \multirow{2}{*}{$1.3\,$M$^{\mathrm{b}}$} & \multirow{2}{*}{$3.7\,$M$^{\mathrm{b}}$}\\
      \cline{2-2}
      of devices & $40\,$M$^{\mathrm{b}}$ & & & \\
      \hline
      \multirow{2}{*}{MNIST} & \multirow{2}{*}{$96.5\,$\%} & $97.02\,$\% & $97.47\,$\% & $97.17\,$\%\\ 
       & & $\pm0.40\,$\% & $\pm0.33\,$\% & $\pm0.64\,$\%\\
      \hline
      \multirow{2}{*}{F-MNIST} & \multirow{2}{*}{$87\,$\%} & $84.87\,$\% & $86.09\,$\% & $87.25\,$\% \\ 
      & & $\pm0.53\,$\% & $\pm0.33\,$\% & $\pm0.35\,$\% \\
      \hline
      \multicolumn{4}{l}{$^{\mathrm{a}}$1T1R configuration\cite{m21}}\\
      \multicolumn{4}{l}{$^{\mathrm{b}}$2T2R configuration}
    \end{tabular}
    \end{center}
    \label{tab:1}
    \vspace{-0.55cm}
\end{table}
\section{Conclusion}
In this work we extended the synaptic metaplasticity training algorithm for BNNs to work with QNNs, implementing it on a mixed analog/digital platform using hafnium oxide memristor crossbars. Our combined software/hardware experiment showed robustness to the main memristors limitations (computational precision, hardware imperfection, and endurance) and achieved equivalent performance to software implementations. We also utilized multi-level programming for compact weight memorization, resulting in a $15\times$ reduction in memory compared to BNNs. Our findings enable QNNs for online synaptic consolidation avoiding catastrophic forgetting. Our work opens up possibilities for developing embedded hardware for continual learning.
\section*{Acknowledgment}
This work was supported by European Research Council consolidator grant DIVERSE (101043854) and by the H2020 MeM-Scales project (871371). It also benefits from a France 2030 government grant managed by the French National Research Agency (ANR-22-PEEL-0010) and the Horizon Europe METASPIN project (101098651).

\end{document}